\documentclass[letterpaper, 10 pt, conference]{ieeeconf}  

\IEEEoverridecommandlockouts                              

\usepackage{graphics} 
\usepackage{epsfig} 
\usepackage{mathptmx} 
\usepackage{times} 
\usepackage{amsmath} 
\usepackage{amssymb}  
\usepackage[ruled,vlined,linesnumbered]{algorithm2e}
\usepackage{hyperref}
\usepackage{bm}
\usepackage{gensymb}
\usepackage{caption}
\usepackage{subcaption}
\usepackage{hyperref}
\usepackage{xcolor}
\usepackage{wrapfig}
\usepackage{algpseudocode}

\renewcommand{\baselinestretch}{0.95}

\title{\LARGE \bf
Reachable Polyhedral Marching (RPM): A Safety Verification Algorithm for Robotic Systems
with Deep Neural Network Components}

\author{Joseph A. Vincent$^{1}$ and Mac Schwager$^{1}$
\thanks{This work was supported by NSF grant 1830402, DARPA grant D18AP00064, NASA grant 80NSSC20M0163, and the Dwight D. Eisenhower Transportation Fellowship. The NASA University Leadership Initiative (grant \#80NSSC20M0163) provided funds to assist the authors with their research, but this article solely reflects the opinions and conclusions of its authors and not any NASA entity. We are grateful for this support.}
\thanks{$^{1}$Department of Aeronautics and Astronautics, Stanford University, Stanford, CA 94305, USA, {\texttt\small \{josephav, schwager\}@stanford.edu}}%
}

\begin{document}

\maketitle
\thispagestyle{empty}
\pagestyle{empty}

\begin{abstract} \label{Abstract}
We present a method for computing exact reachable sets for deep neural networks with rectified linear unit (ReLU) activation.  Our method is well-suited for use in rigorous safety analysis of robotic perception and control systems with deep neural network components. Our algorithm can compute both forward and backward reachable sets for a ReLU network iterated over multiple time steps, as would be found in a perception-action loop in a robotic system.  Our algorithm is unique in that it builds the reachable sets by incrementally enumerating polyhedral cells in the input space, rather than iterating layer-by-layer through the network as in other methods.  If an unsafe cell is found, our algorithm can return this result without completing the full reachability computation, thus giving an anytime property that accelerates safety verification. In addition, our method requires less memory during execution compared to existing methods where memory can be a limiting factor.  We demonstrate our algorithm on safety verification of the ACAS Xu aircraft advisory system. We find unsafe actions many times faster than the fastest existing method and certify no unsafe actions exist in about twice the time of the existing method.  We also compute forward and backward reachable sets for a learned model of pendulum dynamics over a 50 time step horizon in  87s on a laptop computer. Algorithm source code: \url{https://github.com/StanfordMSL/Neural-Network-Reach}.
\end{abstract}

\section{Introduction}\label{Introduction}
\begin{figure}
  \centering
  \includegraphics[width = 0.98\columnwidth]{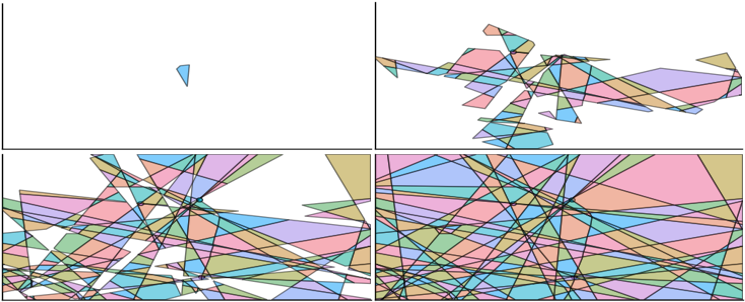}
  \caption{Snapshots of how RPM incrementally enumerates the 2D input space polyhedra for which a random ReLU network is affine, resulting in the explicit piecewise-affine representation. Polyhedron color is random.}
  \label{fig:cell_enum}
\end{figure}

In this paper we present the Reachable Polyhedral Marching (RPM) algorithm for computing forward and backward reachable sets of deep neural networks with rectified linear unit (ReLU) activation.  This is a critical building block to proving safety properties for autonomous systems with learned perception, dynamics, or control components in the loop.  Specifically, given a set of input values, RPM will compute the set of all output values that can be obtained under the ReLU network (the forward reachable set, or image, of the input set).  Similarly, given a set of output values, RPM will compute the set of all input values that can lead to those output values under the ReLU network (the backward reachable set, or preimage, of the output values).  When the ReLU network is part of a dynamical process, as is common in robots with deep learned perception or control components, RPM can compute such reachable sets for multiple time steps into the future or the past without explicitly iterating over each time step.  Figure \ref{fig:cell_enum} shows the incremental nature of how RPM enumerates input space polyhedra over which the ReLU network is affine.

All existing algorithms that compute exact reachable sets iterate through the network layer-by-layer \cite{xiang2017reachable,nnv,yang2020reachability}, which may become intractable if the network is applied iteratively in a feedback loop.  RPM applies a fundamentally different approach to avoid layer-by-layer computation.  Instead, we compute the reachable set one polyhedral cell at a time, where each cell represents a region of the input space over which the network is affine.  Each cell can be computed quickly through a series of linear programs equal to the number of neurons in the network, regardless of width or depth.  Our algorithm computes cells until the explored set of cells fills the desired reachable set. In this way, our method is geometrically similar to fast marching methods in optimal control \cite{HJBFastMarching}, path planning \cite{PlanningFastMarching,FMT}, and graphics \cite{FastMarchingCubes,lei2020analytic}.

We demonstrate RPM in two examples related to the safety verification of dynamical systems with learned components.  First, we compute forward reachable and backward reachable sets for a learned dynamical model of a pendulum over 50 time steps in the future and past, respectively. Furthermore, we compare with a state of the art reachability method \cite{yang2020reachability} in a safety verification problem involving ACAS Xu, a neural network policy for aircraft collision avoidance.  Our algorithm finds unsafe inputs for this network many times faster than \cite{yang2020reachability} and when no unsafe input exists, our algorithm certifies safety approximately 2 times slower.

The paper is organized as follows. We give related work in Section \ref{Related Work} and give background and state the problem in Section \ref{Background}.  Section \ref{PWA} presents our main algorithm, and explains its derivation. In Section \ref{reach} we describe how the RPM algorithm can be used to perform forward and backward reachability over multiple time steps. In Section \ref{Examples} we present the aforementioned examples of RPM.



\section{Related Work} \label{Related Work}
Though the analysis of neural networks is a young field, a broad literature has emerged to address varied questions related to interpretability, verification, and safety. Much work has been dedicated to characterizing the expressive potential of ReLU networks by studying how the number of affine regions scales with network depth and width \cite{bengio,Hanin_2019,understanding}. Other research includes encoding piecewise-affine (PWA) functions as ReLU networks \cite{fem,reverseengineering}, learning deep signed distance functions and extracting level sets \cite{lei2020analytic}, and learning neural network dynamics or controllers that satisfy stability conditions \cite{zico,jin2020neural}, which may be more broadly grouped with correct-by-construction training approaches \cite{art,lincolnlab}.  

Spurred on by pioneering methods such as Reluplex \cite{reluplex}, the field of neural network verification has emerged to address the problem of analyzing properties of neural networks over continuous input sets. A survey of the neural network verification literature is given by \cite{survey}. Reachability approaches are subset of this literature and are especially useful for analysis of learned dynamical systems. Reachability methods can be categorized into overapproximate and exact methods. Overapproximate methods often compute neuron-wise bounds either from interval arithmetic or symbolically \cite{maxsens,fastlip,reluval,neurify,crown}. Optimization based approaches such as (mixed-integer) linear programming are also used to solve for bounds on a reachable output set in \cite{sherlock,alessio1,alessio2,proveable}. Other approaches include modeling the network as a hybrid system \cite{verisig}, abstracting the domain \cite{abstract}, and performing layer-by-layer operations on zonotopes \cite{AI2}. Further, \cite{julian2019reachability} demonstrated how \cite{reluplex, sherlock, reluval, verisig} could be used to perform closed-loop reachability of a dynamical system given a neural network controller.

Exact reachability methods have also been proposed, although to a lesser degree \cite{xiang2017reachable,nnv,yang2020reachability}. These methods generally perform the set operations of affine transformation, intersection/division, and projection layer-by-layer through a ReLU network. Layer-by-layer approaches have also been proposed to solve for the explicit PWA representation of a ReLU network \cite{robinson2020dissecting,hanin2019complexity}. 
Exact methods, unlike overapproximate methods, can compute backward reachable sets.

Our RPM algorithm inherits all advantages of exact methods, but is unique in that it does not iterate layer-by-layer. This can lead to faster verification decisions when finding unsafe inputs. Layer-by-layer methods must compute the entire reachable set, regardless of whether a network violates a safety property. In contrast, if our algorithm encounters a cell with an unsafe input, it can return that result immediately without computing the entire reachable set. Finally, all intermediate polyhedra and affine map matrices of layer-by-layer methods must be stored in memory until the algorithm terminates (since intermediate polyhedra may be split by later neurons). Conversely, since our method computes the true PWA function incrementally, once a polyhedron-affine map pair is computed it can be sent to external memory and only a binary vector (of the neuron activations) needs to be stored to continue the algorithm. This is especially useful because the number of affine regions is hard to estimate before runtime.

\section{Background and Problem Statement} \label{Background}
An $n$-layer feedforward neural network implements a function $F(\textbf{x}) : \mathbb{R}^{k_0} \rightarrow \mathbb{R}^{k_n}$ to give a map from inputs $\mathbf{x}$ to outputs $\textbf{y} = F(\textbf{x})$. Each layer in $F$ is a function $F_i(\textbf{z}_{i-1}) : \mathbb{R}^{k_{i-1}} \rightarrow \mathbb{R}^{k_i}$ where $\textbf{z}_{i} \in \mathbb{R}^{k_i}$ is the hidden layer variable for layer $i$. We assume $\textbf{z}_0 = \textbf{x}$ and $\textbf{z}_n = \textbf{y}$. The function for layer $i$ is
\begin{align}
    F_i(\textbf{z}_{i-1}) = \textbf{z}_{i} = \sigma_i(\hat{\textbf{z}}_{i}) = \sigma_i(\textbf{W}_i \textbf{z}_{i-1} + \textbf{b}_i),
\end{align}
where $\sigma_i$ is the activation function, $\hat{\textbf{z}}_{i}$ is the preactivation value, and $\textbf{W}_i$ and $\textbf{b}_i$ are the weights and biases for layer $i$. We assume $\sigma_{n}$ is an identity map and all hidden layer activations are the rectified linear unit (ReLU)
\begin{align}
    \sigma_i(\hat{\textbf{z}}_{i}) = [max(0,\hat{z}_{i,1}),...,max(0,\hat{z}_{i,k_i})]^\top.
\end{align}

From the above definition we can augment the weights to construct an equivalent neural network that has no bias terms. This equivalent formulation is such that $\tilde{F}(\tilde{\textbf{x}}) : \mathbb{R}^{k_0+1} \rightarrow \mathbb{R}^{k_n}$ and $\tilde{\textbf{x}} = [x_1, x_2, ..., x_n, 1]^\top$. We then have
\begin{subequations}
    \begin{align}
        \tilde{F}_i(\tilde{\textbf{z}}_{i-1}) = \tilde{\textbf{z}}_i = \sigma_i(\tilde{\textbf{W}}_i \tilde{\textbf{z}}_{i-1}) \\
        \tilde{\textbf{W}}_i = 
        \begin{bmatrix}
            \textbf{W}_i & \textbf{b}_i \\
            \textbf{0}^\top & 1
        \end{bmatrix} \\
        \tilde{\textbf{W}}_n = 
    \begin{bmatrix}
        \textbf{W}_n & \textbf{b}_n
    \end{bmatrix}.
    \end{align}
\end{subequations}

It is well-known that the function implemented by a ReLU network is continuous and PWA \cite{bengio,Hanin_2019,understanding}.  We mathematically characterize PWA functions below.
\newtheorem{definition}{Definition}
\begin{definition}[Polyhedral Complex]
A polyhedral complex $\mathcal{C}$ is a finite set of convex polyhedra where every polyhedron contains its faces and the intersection between two polyhedra is either a face of both $c_1$ and $c_2$ or empty. We henceforth refer to convex polyhedra simply as polyhedra.
\end{definition}

A PWA function is a function $M(\textbf{x}) : \mathbb{R}^{n} \rightarrow \mathbb{R}^{m}$ where
\begin{align}
    M(\textbf{x}) = \textbf{C}_i \textbf{x} + \textbf{d}_i \ \ \ \forall \textbf{x} \in c_i \ \ \ \forall c_i \in \mathcal{C} \label{eq:PWA}
\end{align}
for $\textbf{C}_i \in \mathbb{R}^{m \times n}$, $\textbf{d}_i \in \mathbb{R}^{m}$, and $c_i$ elements of a polyhedral complex $\mathcal{C}$. Here we consider only continuous PWA functions. Equation (\ref{eq:PWA}) is the \textit{explicit} PWA representation.

The forward reachability problem is to solve for the image of a specified input set under the neural network map
\begin{align}
    \mathcal{Y} = \{F(\textbf{x}) \vert \textbf{x} \in \mathcal{X} \}.
\end{align}
Likewise, the backward reachability problem is to solve for the preimage of a specified output set
\begin{align}
    \mathcal{X} = \{x \vert F(\textbf{x}) \in \mathcal{Y} \}.
\end{align} 

Our algorithm transforms a ReLU network into an explicit PWA representation by computing a set of polyhedra ($c_i$) and associated affine maps $(\textbf{C}_i, \textbf{d}_i)$.  There exist efficient methods for computing forward and backward reachable sets of polyhedral sets under PWA maps. Accordingly, we restrict our focus to these problems. 


\section{Explicit PWA Representation} \label{PWA}
First, we seek to construct the explicit PWA representation of a ReLU network. Our method constructs a PWA function for a ReLU network by enumerating each polyhedral cell and its associated affine map. In Sec.~\ref{Cells} we show how cells and affine maps are computed from the \textit{activation pattern} of a ReLU network. In Sec.~\ref{Essential} we show how cell representations are reduced to a minimal form, which is used in Sec.~\ref{Neighbors} to determine neighboring polyhedra given a current polyhedron. This leads to a recursive procedure in which we explore an expanding front of polyhedra, ultimately giving the explicit PWA representation, as explained in Sec.~\ref{Cell Enumeration}.

\subsection{Determining Polyhedral Cells from Activation Patterns} \label{Cells}
For a given input to a ReLU network, every hidden neuron has an associated binary neuron activation of zero or one corresponding to whether the preactivation value was nonpositive or positive, respectively\footnote{This choice is arbitrary and some others use the opposite convention.}. We refer to the vector of all neuron activations as the \textit{activation pattern} of the network for a given input. For neuron $j$ in layer $i$,
\begin{subequations}
    \begin{align}
        AP_{ij}(\textbf{x}) = \begin{cases}
                                 1\ \  \hat{z}_{ij} > 0 \\    
                                 0\ \  \hat{z}_{ij} \le 0 
                              \end{cases} \label{eq:AP_1} \\
        AP_i(\textbf{x}) = [ AP_{i,1}(\textbf{x}), ..., AP_{i,k_i}(\textbf{x})  ]^\top \\
        AP(\textbf{x}) = (AP_1(\textbf{x}), ..., AP_{n-1}(\textbf{x})). \label{eq:AP}
    \end{align}
\end{subequations}
For a network with $N$ hidden neurons there are $2^N$ possible combinations of neuron activations, however not all are realizable by the network. Practically, the number of activation patterns is better approximated by $N^{k_0}$ \cite{hanin2019deep}.

To construct an explicit PWA representation we first want to characterize the set of inputs $c$ for which the network reduces to a single affine map. For a fixed activation pattern the ReLU network simplifies to the affine map $\textbf{C} \textbf{x} + \textbf{d}$ where
\begin{subequations}
    \begin{align}
        \begin{bmatrix}
            \textbf{C} & \textbf{d}
        \end{bmatrix} = 
        \tilde{\textbf{W}}_n \prod_{i=1}^{n-1} \tilde{\textbf{W}}^c_i \label{eq:linear} \\
        \tilde{\textbf{W}}^c_i(\textbf{x}) = diag(AP_{i}(\textbf{x}))\tilde{\textbf{W}}_i, \label{eq:diag}
    \end{align}
\end{subequations}
and $diag(\cdot)$ diagonalizes a vector to a matrix. Equation (\ref{eq:diag}) equivalently sets row $j$ of $\tilde{\textbf{W}}_i$ to zero if $AP_{ij} = 0$.

The activation pattern only changes if some $\hat{z}_{ij}$ switches from positive to nonpositive (or vice-versa), $c$ is thus given by a set of linear constraints, one for each neuron in the network.  The constraint for neuron $j$ in layer $i$ is
\begin{align}
    \begin{cases}
          \textbf{a}_{ij}^{\top} \textbf{x} \ge b_{ij}\ \ \text{if}\  AP_{ij} = 1 \\    
          \textbf{a}_{ij}^{\top} \textbf{x} \le b_{ij}\ \ \text{if}\  AP_{ij} = 0
      \end{cases} \label{eq:constraints}
\end{align}
where
\begin{align}
    \begin{bmatrix}
        \textbf{a}_{ij} & -b_{ij}
    \end{bmatrix} =  
    \tilde{\textbf{W}}_i[j,:] \prod_{l=1}^{i-1} \tilde{\textbf{W}}_l, \label{eq:ab}
\end{align}
and $\tilde{\textbf{W}}_i[j,:]$ denotes the $j^{th}$ row of $\tilde{\textbf{W}}_i$. Equation (\ref{eq:ab}) is similar to (\ref{eq:linear}) but instead simply gives the affine map parameters for a single neuron output. Strict inequalities are not present in (\ref{eq:constraints}) because the affine map $\textbf{Cx} + \textbf{d}$ holds over the interior and boundary of the set where the activation pattern is constant. Since $c$ is an intersection of halfspaces, the set is a polyhedron. A similar formulation is given in \cite{lei2020analytic} for finding the surface of an object for graphics rendering. Note that it is not uncommon for neurons to have the degenerate linear constraint $\textbf{0}^{\top} \textbf{x} \le 0$ which is satisfied for all $\textbf{x} \in c$, and also for multiple different neurons in the network to have equivalent constraints (perhaps scaled arbitrarily). To address these edge cases we introduce Definitions \ref{def: hrep}, \ref{def: duplicate}, \ref{def: redundant} before describing our fast marching method below.  

\begin{definition}[H-representation]
The halfspace representation (H-representation) of a polyhedron is  
\begin{align}
    P_H = \{\textbf{x} \in \mathbb{R}^d \vert \textbf{A} \textbf{x} \le \textbf{b} \} \label{eq:hrep}
\end{align}
where $\textbf{A}$ an $m \times d$ matrix and $\textbf{b}$ an $m$-dimensional vector. Equality constraints may also be included if the set $P_H$ is of lower dimension than the ambient dimension. \label{def: hrep}
\end{definition}

\begin{definition}[Duplicate Constraints]
A constraint $\textbf{a}_j^{\top} \textbf{x} \le b_j$ is duplicate if there exists a scalar $\alpha$ and a prior constraint $\textbf{a}_i^{\top} \textbf{x} \le b_i$ where $i<j$ such that $\alpha [\textbf{a}_j\ b_j] = [\textbf{a}_i\ b_i]$ where $\alpha > 0$. \label{def: duplicate}
\end{definition}
\begin{definition}[Redundant \& Essential Constraints]
A constraint $\textbf{a}_i^{\top} \textbf{x} \le b_i$ is redundant if the feasible set does not change upon its removal or if it is a duplicate constraint. An essential constraint is one which is not redundant. \label{def: redundant}
\end{definition}

We use (\ref{eq:constraints}) to construct an H-representation of a cell $c$ by enumerating all constraints associated with hidden layer neurons.  The resulting $\textbf{a}_{ij}$ vectors and $b_{ij}$ scalars are collected into ($\textbf{A}$, $\textbf{b}$) to give the desired H-representation from (\ref{eq:hrep}).  For an arbitrary ReLU network in an arbitrary activation pattern, we find that the resulting H-representation often has redundant constraints, which need to be removed to give a minimal representation of the polyhedron.  



\subsection{Finding Essential Constraints} \label{Essential}
We first normalize all constraints, remove any duplicates, and consider the resulting  H-representation $\textbf{A}\textbf{x} \le \textbf{b}$. To determine if the $i$th constraint is essential or redundant, define a new set of constraints with the $i$th constraint removed,
\begin{subequations}
    \begin{align}
        \tilde{\textbf{A}} = [\textbf{a}_1\ ...\ \textbf{a}_{i-1}\ \textbf{a}_{i+1}\ ...\  \textbf{a}_m]^\top \\
         \tilde{\textbf{b}} = [b_1\ ...\ b_{i-1}\ b_{i+1}\ ...\  b_m]^\top,
    \end{align}
\end{subequations}

and solve the linear program
\begin{subequations}
    \begin{align}
        \max_{\textbf{x}} \quad & \textbf{a}^{\top}_i \textbf{x}\\
        \textrm{subject to} \quad & \tilde{\textbf{A}}\textbf{x} \le \tilde{\textbf{b}}.
    \end{align}
    If the optimal objective value is less than or equal to $b_i$, constraint $i$ is redundant.
\end{subequations}
In the worst case, a single LP must be solved for each constraint to determine whether it is essential or redundant. However, heuristics exist to avoid this worst case complexity. We use the bounding box heuristic to quickly find a subset of redundant constraints \cite{morari}. Empirically, this results in identifying about $90\%$ of the redundant constraints. We find that other heuristics beyond the bounding box method do not improve performance. Our implementation uses the GLPK open-source LP solver and the JuMP optimization package \cite{DunningHuchetteLubin2017}.


\subsection{Determining Neighboring Activation Patterns} \label{Neighbors}
Given a cell $c$ in the input space of a ReLU network, we are interested in finding the activation pattern corresponding to a neighboring cell, $c'$. We can then determine the H-representation of $c'$ using the procedure outlined above. The challenge in finding the activation pattern for $c'$ is in determining which individual neuron activations ($AP_{ij}$ for some $ij$) must be flipped when $\textbf{x}$ transitions from $c$ to $c'$. 

The boundary between $c$ and $c'$ ($c \cap c'$) is defined by a linear constraint of $c$. We refer to this as the \textit{neighbor constraint}. Intuitively, to generate $AP^{c'}$ one ought to simply flip the activation of the neuron defining this neighbor constraint (and all neurons defining duplicate constraints). However, this intuitive procedure is incomplete because it does not correctly deal with degenerate constraints $\textbf{0}^{\top} \textbf{x} \le 0 \ \forall \textbf{x} \in c$, which are common in practice.  A degenerate constraint holds everywhere over $c$, but it may turn into a non-degenerate constraint for $c'$ due to a neuron activation in an earlier layer being flipped when moving to cell $c'$. Conversely, a non-degenerate constraint in $c$ may turn into a degenerate one when passing to $c'$.  To define a correct procedure for flipping neuron activations we introduce a taxonomy for classifying neurons based on their associated linear constraints.

\begin{figure}
     \centering
     \includegraphics[width=0.95\columnwidth]{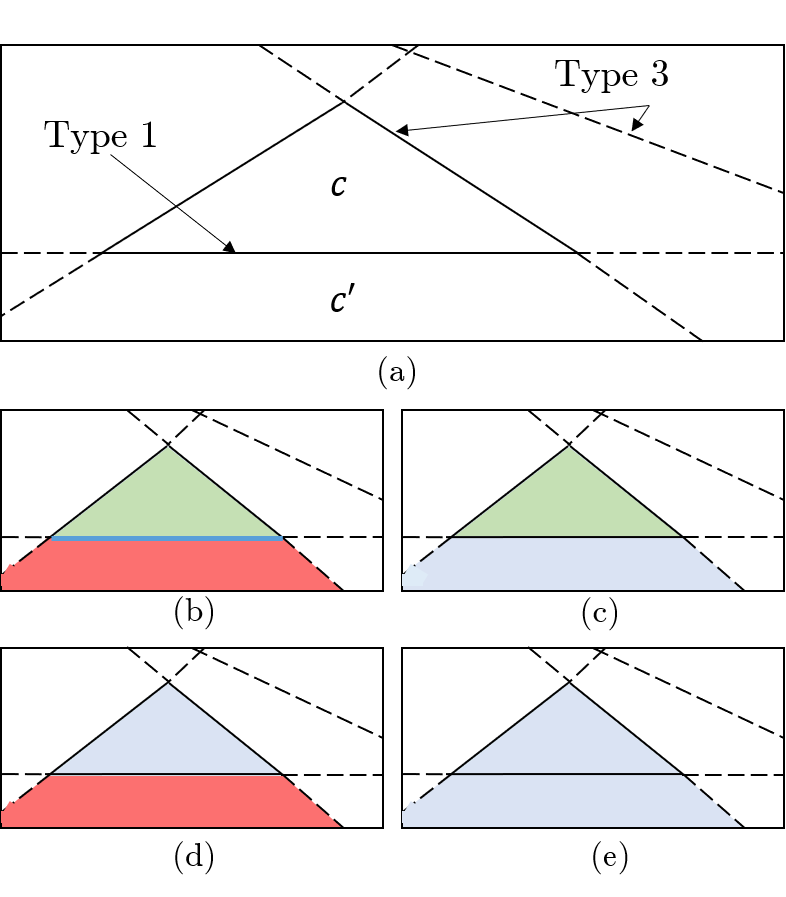}
     \caption{(a) Example $c$ and $c'$ with Type 1 and Type 3 constraints labeled. For (b)-(e), regions in green denote where the nonlinear constraint is strictly satisfied, regions in blue denote where the nonlinear constraint function evaluates to zero, and regions in red denote where the nonlinear constraint is violated. (b) Type 1 constraint where the neuron activation must flip when going from $c$ to $c'$. (c) Type 1 constraint where the neuron activation is set to zero when going from $c$ to $c'$. (d) Type 2 constraint where the neuron activation must flip when going from $c$ to $c'$. (e) Type 2 constraint where the neuron activation remains zero when going from $c$ to $c'$.}
     \label{fig:constraints}
\end{figure}

\begin{definition}[Type 1, 2, 3 neurons] \label{types}
When identifying a neighboring cell $c'$ from a current cell $c$, neurons whose linear constraints in $c$ define $c \cap c'$ are called \textit{Type 1}. neurons with degenerate linear constraints in $c$, $\textbf{0}^{\top} \textbf{x} \le 0 \ \forall \textbf{x} \in c$, are called \textit{Type 2}. All other neurons are \textit{Type 3}. 
\end{definition}
Constraints labeled by type are illustrated in Figure \ref{fig:constraints}a. 

Though not stated explicitly, the linear constraint parameters from (\ref{eq:constraints}) are (nonlinear) functions of the input. Given Definition \ref{types}, for inputs $\textbf{x} \in c \cap c'$,
\begin{subequations}
    \begin{align}
        \textbf{a}_{ij}^{\top}(\textbf{x}) \textbf{x} = b_{ij}(\textbf{x})\ \ \text{if}\  ij \in \text{Type 1} \\
        \textbf{a}_{ij}^{\top}(\textbf{x}) \textbf{x} = b_{ij}(\textbf{x})\ \ \text{if}\  ij \in \text{Type 2} \\
        \begin{cases}
            \textbf{a}_{ij}^{\top}(\textbf{x}) \textbf{x} > b_{ij}(\textbf{x})\ \ \text{if}\  ij \in \text{Type 3}\ \text{and}\ AP_{ij} = 1 \\
            \textbf{a}_{ij}^{\top}(\textbf{x}) \textbf{x} < b_{ij}(\textbf{x})\ \ \text{if}\  ij \in \text{Type 3}\ \text{and}\ AP_{ij} = 0
        \end{cases}. \label{eq:T3}
    \end{align}
\end{subequations}
The function $\textbf{a}_{ij}^{\top}(\textbf{x}) \textbf{x} - b_{ij}(\textbf{x})$ is continuous because it is simply the function for computing the preactivation value $\hat{z}_{ij}$. From this continuity, we know that the strict inequality of (\ref{eq:T3}) holds for all inputs in $c'$ as well. Therefore, the activation of a Type 3 neuron is not flipped. 

Conversely, inputs in $c'$ may be either feasible or infeasible for Type 1 and Type 2 constraints. Figure \ref{fig:constraints}b-c illustrates the possible feasible and infeasible regions of a Type 1 constraint. Figure \ref{fig:constraints}d-e illustrates the possible feasible and infeasible regions of a Type 2 constraint. It follows that the procedure to generate a neighboring activation pattern is to apply (\ref{eq:constraints}) layer-by-layer updating the activation pattern in place, where Type 1 and Type 2 neuron activations are set to zero if their new linear constraints are $\textbf{0}^{\top} \textbf{x} \le 0$, otherwise, they are flipped. This procedure is formally defined in Algorithm \ref{Algo:neighbor_ap}.  We note that applying this neuron flipping algorithm to identify a neighboring cell $c'$ is exceedingly fast, even for networks with 10,000s of neurons, as we only require a single logical check at each neuron. 
\begin{algorithm}[!h]
\SetAlgoLined
\DontPrintSemicolon
 \KwIn{$AP^{c}$, $Type\ 1$, $Type\ 2$,  $(W_0, \ldots, W_N$)}
 \KwOut{$AP^{c'}$}
 $AP^{c'} \leftarrow  AP^{c}$ \algorithmiccomment{Initialize new AP as old AP} \\
 
\For(\algorithmiccomment{For each neuron}){$i \in [1,...,n-1]$, $j \in [1,...k_i]$}{ 
\If{$ij \in Type\ 1\ or\ ij \in Type\ 2$}{
$\textbf{a}_{ij},\ b_{ij} \leftarrow$ Equation \ref{eq:ab} \algorithmiccomment{$AP^{c'}_{ij}$ hyperplane} \\
\uIf{$[\textbf{a}_{ij}\ b_{ij}] = \textbf{0}$}{$AP^{c'}_{ij} \leftarrow 0$}
\Else{$AP^{c'}_{ij} \leftarrow \neg AP^{c'}_{ij} $}
}
}
 
\Return{$AP^{c'}$}
\caption{Neighboring Activation Pattern}
\label{Algo:neighbor_ap}
\end{algorithm}

\subsection{Reachable Polyhedral Marching} \label{Cell Enumeration}
From (\ref{eq:constraints}) and Algorithm \ref{Algo:neighbor_ap}, we define our main algorithm, Reachable Polyhedral Marching (RPM) in Algorithm \ref{Algo:cell_enumeration} for explicitly enumerating all polyhedral cells in the input space of a ReLU network, resulting in the explicit PWA representation. First, we start with an initial point in the input space and evaluate the network to find the activation pattern. From this the H-representation is found using (\ref{eq:constraints}). Essential constraints are then identified as described in Section \ref{Essential}. For each essential constraint we generate a neighboring activation pattern using Algorithm \ref{Algo:neighbor_ap}. The process then repeats with each new neighboring activation pattern being added to a working set. A neighbor activation pattern only gets added to the working set if it has not already been visited and it is not already in the working set. For a given starting cell each neighbor cell is enumerated, and since each is connected to another, Algorithm \ref{Algo:cell_enumeration} is guaranteed to enumerate every cell in the input space. Figure \ref{fig:cell_enum} shows the result of applying Algorithm \ref{Algo:cell_enumeration} to a randomly initialized ReLU network.
\begin{algorithm}
\SetAlgoLined
\DontPrintSemicolon
 \KwIn{$AP^{c_0}$, $(W_0, \ldots, W_N$)}
 \KwOut{Explicit PWA representation}
 $input\ cells$ = $\emptyset$;  $visited\ cells$ = $\emptyset$ \\
 $working\ set$ = \{$AP^{c_0}$\} \\
 \While{$working\ set \neq \emptyset$}{
 $AP^{c} \leftarrow$ pop element off of $working\ set$ \\
 $\textbf{C}^{c},\ \textbf{d}^{c} \leftarrow$ Equation \ref{eq:linear} \algorithmiccomment{Retrieve affine map} \\
 $H^{c}_{rep} \leftarrow$ Equation \ref{eq:constraints} \algorithmiccomment{Retrieve H-representation} \\
 $H^{c}_{rep} \leftarrow$ remove redundant $H^{c}_{rep}$ constraints \\
 push ($\textbf{C}^{c},\ \textbf{d}^{c},\ H^{c}_{rep}$) onto $input\ cells$ \\
 \For{$\textbf{a}_k, b_k \in H^{c}_{rep}$}{
 $AP^{c'} \leftarrow$ Algorithm \ref{Algo:neighbor_ap} \\
 \If{$AP^{c'} \notin visited\ cells \cup working\ set$}{
 push $AP^{c'}$ onto $working\ set$ \\
 }
 }
 push $AP^{c}$ onto $visited\ cells$ \\
 }
 \Return{input cells}
 \caption{Reachable Polyhedral Marching}
 \label{Algo:cell_enumeration}
\end{algorithm}

\section{Reachability} \label{reach}
\subsection{Forward Reachability} \label{Forward}
The forward reachable set of a PWA function over some input set is simply the union of forward reachable sets of each individual polyhedron under its associated affine map. The image of a polyhedron under an affine map is
\begin{align}
    P_{out} = \{\textbf{y} \vert \textbf{y}=\textbf{Cx}+\textbf{d}, \textbf{Ax} \le \textbf{b}\}. \label{eq:forward reach}
\end{align}
For $\textbf{C}$ invertible, the H-representation of the image is
\begin{align}
    P_{out} = \{\textbf{y} \vert \textbf{AC}^{-1}\textbf{y} \le \textbf{b} + \textbf{AC}^{-1} \textbf{d}\}. \label{eq:image}
\end{align}
In the case the affine map is not invertible, more general polyhedral projection methods such as block elimination, Fourier-Motzkin elimination, or parametric linear programming can compute the H-representation of the image \cite{cdd_manual, mpLP}. Our implementation uses the block elimination projection in the case of a non-invertible affine map \cite{cdd}.

Our RPM algorithm is used to perform forward reachability as follows. We first specify a polyhedral input set whose image through the ReLU network we want to compute.  This is a set over which to perform the RPM algorithm. For each activation pattern $AP^{c}$ we also compute the image of $H^{c}_{rep}$ under the map $\textbf{C}^{c}\textbf{x} + \textbf{d}^{c}$. To do this, an additional line is introduced between lines 9 and 10 of Algorithm \ref{Algo:cell_enumeration}
\begin{align}
    H^{c}_{rep,\ forward} \leftarrow project(H^{c}_{rep}, \textbf{C}^{c}, \textbf{d}^{c})
\end{align}
where $project(\cdot, \cdot, \cdot)$ applies (\ref{eq:image}) if $\textbf{C}^{c}$ invertible and block elimination otherwise.

\subsection{Backward Reachability} \label{Backward}
The preimage of a polyhedron under an affine map is
\begin{subequations}
    \begin{align}
        P_{in} = \{\textbf{x} \vert \textbf{y}=\textbf{Cx}+\textbf{d}, \textbf{Ay} \le \textbf{b}\} \\
        P_{in} = \{\textbf{x} \vert \textbf{ACx} \le \textbf{b}-\textbf{Ad}\}. \label{eq:backward reach}
    \end{align}
\end{subequations}
Like forward reachability, performing backward reachability only requires a small modification to Algorithm \ref{Algo:cell_enumeration}. We first specify a polyhedral output set whose preimage we would like to compute.  For each activation pattern $AP^{c}$, we also compute the intersection of $H^{c}_{rep}$ with the preimage of the given output set under the map $\textbf{C}^{c}\textbf{x} + \textbf{d}^{c}$. Two additional lines are thus introduced between lines 9 and 10 of Algorithm \ref{Algo:cell_enumeration}
\begin{subequations}
    \begin{align}
    P^{c}_{in} \leftarrow \text{Equation}\ \ref{eq:backward reach} \label{eq:p_in} \\
    H^{c}_{rep,\ backward} \leftarrow H^{c}_{rep} \cap P^{c}_{in}. \label{eq:h_back}
    \end{align}
\end{subequations}
Multiple backward reachable sets can be solved for simultaneously at the added cost of repeating (\ref{eq:p_in}) and (\ref{eq:h_back}) for each output set argument to Algorithm \ref{Algo:cell_enumeration}.

Finally, we address the issue of finding forward and backward reachable sets iterated over multiple time steps.  For this, we note that a ReLU network that is applied iteratively over $T$ timesteps, $\textbf{x}_{t+1} = F(\textbf{x}_t)$ for $t = 0, \ldots, T-1$, is mathematically equivalent to a single ReLU network consisting of $T$ copies of the original network concatenated end to end, $\textbf{x}_T = F_T(\textbf{x}_0) := F \circ \cdots \circ F(\textbf{x}_0)$.  If the original network $F(\textbf{x})$ has $N$ neurons and $L$ layers, the equivalent multi-time step network $F_T(\textbf{x})$ has $TN$ neurons and $TL$ layers.  We simply perform RPM for forward or backward reachability on the equivalent multi-time step network $F_T(\textbf{x})$.

\section{Examples} \label{Examples}
All examples are run on a 2013 Dell Latitude E6430s laptop with Intel Core i7 3GHz processor and 16GB of RAM. The cell coloring in plots is random.
\subsection{Damped Pendulum Example} \label{Pendulum}

\begin{figure}[!h]
    \centering
    \includegraphics[width = 0.6\columnwidth]{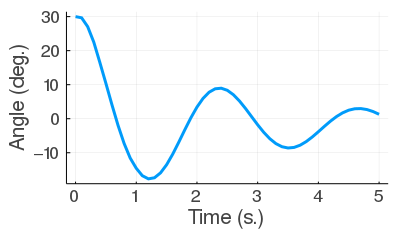}
    \caption{50 time-step trajectory from $\bm{x}_0 = [30, 0]^\top$}
    \label{fig:damped}
\end{figure}

The forward and backward reachability algorithms can be applied to discrete-time dynamical systems represented as ReLU networks. In this example we analyze a ReLU network that approximates the discrete-time dynamics of a damped pendulum. The function learned by the network is 
\begin{align}
    \bm{x}_{t+1} = f_{nn}(\bm{x}_t)
    \label{pendulum}
\end{align}
where $\bm{x} = [\theta,  \dot{\theta}]^\top$. The time-step used is $0.1$ seconds. The learned dynamics function $f_{nn}$ is a ReLU network with a single hidden layer of 12 neurons. An example trajectory of the learned system is shown in Figure \ref{fig:damped}. We can compose multiple copies of $f_{nn}$ together to get a neural network which outputs an arbitrary future state. For instance, composing the network with itself 50 times results in a final network with 600 neurons and output $\bm{x}_{t+50}$. 

\begin{figure*}
     \centering
     \begin{subfigure}[b]{0.95\columnwidth}
         \centering
         \includegraphics[width=\textwidth]{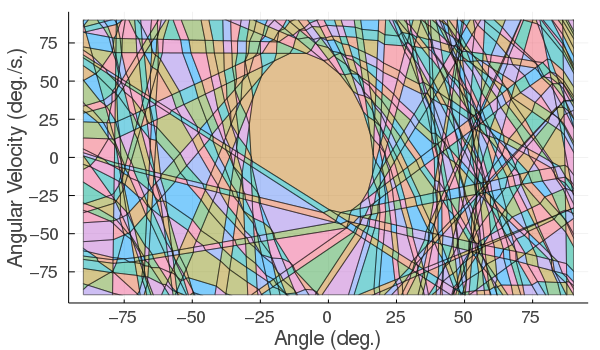}
         \caption{Input cells.}
         \label{fig:pend_in}
     \end{subfigure}
     \hfill
     \begin{subfigure}[b]{0.95\columnwidth}
         \centering
         \includegraphics[width=\textwidth]{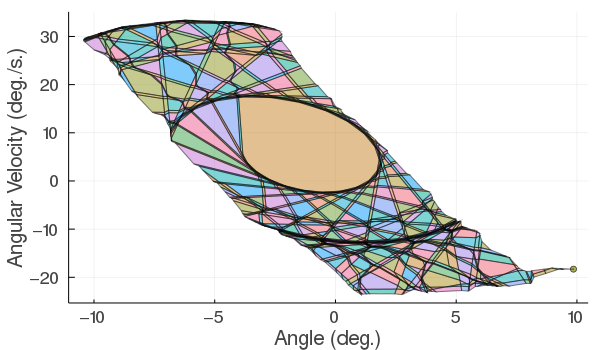}
         \caption{Output cells.}
         \label{fig:pend_out}
     \end{subfigure}
        \caption{(a) Input cells of the 50 time-step pendulum network. (b) image of input cells under the neural network map.}
        \label{fig:forward}
\end{figure*}

Figure \ref{fig:damped} suggests that states of the learned system tend toward the origin. We can use the forward reachability algorithm to determine where \textit{all} states within some set of initial conditions lead to over some time interval. In Figure \ref{fig:pend_in} we show the result of applying the cell enumeration algorithm and the forward reachability algorithm to the 50 time step system for initial conditions $-90 \degree \le \theta \le 90 \degree$ and $-90 \degree / s \le \dot{\theta} \le 90 \degree / s$. The resulting number of regions is 2781 and the forward reachability version of the RPM algorithm took 87s to complete. It is clear from Figure \ref{fig:pend_in} that all initial conditions are mapped to a subset of the initial input set, proving that the original input set is forward invariant over 50 time step increments, and indicating that trajectories tend to approach the origin over time.

\begin{figure}
    \centering
    \includegraphics[width = 0.9\columnwidth]{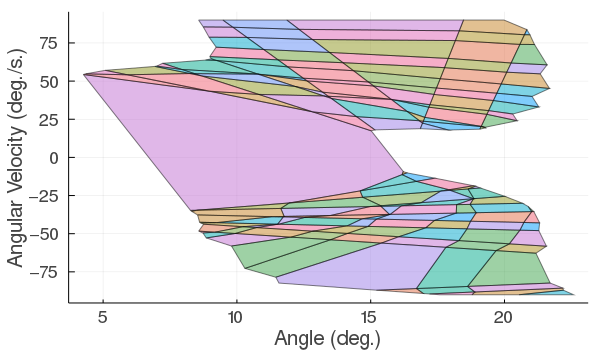}
    \caption{Set of initial states that lead to a neighborhood around the origin after 50 time steps for the pendulum network.}
    \label{fig:back_reach}
\end{figure}

Further, we perform backward reachability to find the set of inputs that map to a small neighborhood around the origin after 50 time steps. Figure \ref{fig:back_reach} shows the inputs that map to the output set $-5 \degree \le \theta \le 5 \degree$ and $-2 \degree / s \le \dot{\theta} \le 2 \degree / s$ after 50 time steps. Interestingly, only initial conditions with positive angles map to the target set. In the true dynamics, the initial angles would be symmetric about zero. This analysis can thus help us identify undesirable modeling artifacts to inform retraining or control of the model.

\subsection{Aircraft Collision Avoidance Example} \label{ACAS}
The ACAS Xu networks are 45 distinct policy networks designed to issue advisory warnings to avoid mid-air collisions for unmanned aircarft. Each network has five inputs, five outputs (advisories), and 300 neurons. Inputs are relative distance, angles, and speeds. See \cite{acas} for more details. 

The appendix of \cite{reluplex} lists ten safety properties the networks should satisfy. Here we consider Property 3. This property is satisfied if the network never outputs a ``Clear of Conflict" advisory when the intruder is directly ahead and moving towards the ownship. We use the backward reachability algorithm to verify whether Property 3 is satisfied. A nonempty backward reachable set implies that some allowable inputs will map to unsafe outputs and the safety property is not satisfied. Table \ref{table:verification} shows the results of verifying Property 3 for four of the networks. A comparison is given against the fastest existing exact reachability method based on the face lattice representation \cite{yang2020reachability} as well as the Reluplex and Marabou verification algorithms \cite{reluplex,marabou}. Each algorithm was run on a single core. We note that a parallelized algorithm is also provided for the face lattice approach \cite{yang2020reachability}, and our RPM algorithm is also amenable to a parallelized implementation to be explored in future work. 

The results in Table \ref{table:verification} show the advantage of our proposed method when verifying network properties that are found to not hold. Our RPM algorithm is anytime, in the sense that once an unsafe input polyhedron is found, the algorithm can terminate without completing the full reachability computation. This is in contrast to all existing exact reachability methods that proceed layer-by-layer through the network.  They must solve the entire backward reachability problem to conclude any verification result, whether safe or unsafe.

\begin{table}
\centering
\caption{ACAS Xu Property 3 Verification Results}
 \begin{tabular}{|c| c| c| c| c| c|} 
 \hline
 Network & Result & \begin{tabular}{@{}c@{}}Time (s) \\ RPM\end{tabular} & \begin{tabular}{@{}c@{}}Time (s) \\ Face Lattice\end{tabular} & \begin{tabular}{@{}c@{}}Time (s) \\ Reluplex\end{tabular} & \begin{tabular}{@{}c@{}}Time (s) \\ Marabou\end{tabular} \\ [0.5ex] 
 \hline
 $N_{1,7}$ & Unsafe & \textbf{0.01} & 6.66 & 2.15 & 0.70 \\ 
 \hline
 $N_{1,8}$ & Unsafe & \textbf{0.01} & 5.45 & 4.32 & 1.49 \\
 \hline
 $N_{3,8}$ & Safe & 19.72 & \textbf{9.35} & 231.28 & 40.13 \\
 \hline
 $N_{5,6}$ & Safe & 31.92 & \textbf{17.28} & 366.39 & 54.07 \\
 \hline
\end{tabular}
\label{table:verification}
\end{table}

\section{Conclusion}
We proposed the Reachable Polyhedral Marching 
(RPM) algorithm to efficiently construct the exact PWA representation of a ReLU network. RPM computes an explicit PWA function representation for a given ReLU network.  This PWA function can then be used to quickly find forward and backward reachable sets over multiple time steps. Solving for multiple backward reachable sets can be done simultaneously at little added cost. RPM is shown to be especially fast when searching for the existence of unsafe inputs during verification.  In the future, we will investigate parallel implementations of RPM to further improve computational speed.





\bibliographystyle{./IEEEtran} 
\bibliography{./IEEEabrv,./IEEEexample}

\end{document}